\documentclass{article}
\usepackage{amssymb, graphicx, amsmath, amsthm,nicematrix}
\usepackage{tikz}
\usepackage[colorlinks=true,linkcolor=blue,citecolor=blue]{hyperref}
\usetikzlibrary{arrows,matrix,positioning}

\theoremstyle{definition}
\newtheorem{theo}{Theorem}
\newtheorem{defi}[theo]{Definition}
\newtheorem{exam}[theo]{Example}

\newtheorem{rema}[theo]{Remark}

\title{Note on computational complexity \\ of the Gromov-Wasserstein distance}
\author{Natalia Kravtsova$^1$}
\date{%
    $^1$kravtsova.2@osu.edu
}

\begin{document}

\maketitle

\begin{abstract} 
    %This note addresses the property frequently mentioned in the literature that the Gromov-Wasserstein distance . 
    This note addresses computational difficulty of the Gromov-Wasserstein distance frequently mentioned in the literature. 
    We provide details on the structure of the Gromov-Wasserstein optimization problem that show its non-convex quadratic nature for any instance of an input data. We further illustrate the non-convexity of the problem with several explicit examples.
    
    %We provide the details on the non-convex nature of the Gromov-Wasserstein distance optimization problem that imply NP-hardness of the Gromov-Wasserstein distance between finite spaces for any instance of an input data. We further illustrate the non-convexity of the problem with several explicit examples.

%% If you want to inform the reader of the paper about your
%% supplementary material, you can refer to it this way. The file
%% itself should be placed in a directory called Attach.

\end{abstract}

The Gromov-Wasserstein (GW) distance defined in \cite{Memoli2007, Memoli} accepts a pair of metric-measure spaces and outputs a distance between them\footnote{up to measure-preserving isometries, which makes GW a metric on the equivalence classes of metric-measure spaces, see Theorem 5.1(a) of \cite{Memoli}.}. This metric property makes GW distance highly applicable for comparison of objects, in particular objects that lie in different spaces,  - a desirable task in scientific and machine learning applications. Scientific applications include aligning density maps in cryo-EM \cite{Riahi}, comparing molecular structures in chemistry \cite{Peyre}, inferring cell-cell communications \cite{Cang} and   
aligning data from different experiments \cite{Demetci} in biology, and classifying biomedical time series \cite{Kravtsova}, to name a few. Machine learning applications pertain to generative modeling \cite{Bunne}, reinforcement learning \cite{Fickinger}, and graph partitioning and clustering \cite{ChowdhuryNeedham, Vayer, Xu2022}, among others. 

In order to perform GW based object comparison, one needs to compute a GW distance, a constrained minimization problem frequently mentioned to be computationally difficult due to its non-convexity (see, e.g., discussion in Section 7 of \cite{Memoli}, Section 1.1 of \cite{Peyre}, Section 1 of \cite{Scetbon}, and/or Section 3 of \cite{Chen}). This note provides the details on the structure of the GW optimization problem for finite spaces - the case most commonly encountered in real life applications - that demonstrate the non-convex quadratic nature of the GW distance on finite spaces. 

A general instance of non-convex quadratic programming is NP-hard \cite{Vavasis2009}. Furthermore, restrictions to several specific subclasses of non-convex quadratic programs were proved to be NP-hard as well. Among these are non-convex \textit{quadratic knapsack} problem \cite{pardalos_knapsack} (NP-hardness established using \textit{maximum clique} problem\footnote{given a graph $G$, find a maximal size subset of vertices which are all connected}) and quadratic problem with a single negative eigenvalue \cite{Pardalos} (NP-hardness established using \textit{clique} problem\footnote{given a graph $G$ and a positive integer $k$, decide whether $G$ has a subset of size $k$ of vertices that are all connected.}). The former has specific constraints (the optimization variable is constrained to be a probability distribution), while the latter has specific objective structure (a single negative eigenvalue). 
At the same time, there are subclasses of non-convex quadratic programs that are shown to be solvable in polynomial time, e.g. a program with non-negativity constraints and a fixed (in terms of the length of optimization variable) number of negative eigenvalues \cite{pardalospolynomial}. Non-convex quadratic nature of GW distance that we illustrate here suggests that this subclass of non-convex quadratic programs could be NP-hard in its general instance. We are not aware of the proof that would utilize the reduction from a known NP-hard problem to establish NP-hardness of the GW distance similarly to establishing NP-hardness of the subclasses mentioned above.

To solve the GW problem in practice, two main approaches are currently taken in the literature, none of which guarantees a global minimum. The first approach is to employ the \textit{conditional gradient} method (also known as \textit{Frank-Wolfe}), which converges to a stationary point of the objective function that may not be a global minimum in the non-convex case (see Sections 2.1 and  2.2.2 of \cite{Bertsekas}). Conditional gradient for GW distance is implemented in the Python toolbox \cite{POT} and used for variants of GW problem in \cite{Chapel, Bai}. The second approach proposed in \cite{Peyre} is to regularize the objective by adding an entropy term (see equation (7) of \cite{Peyre} and discussion in Section 10.6.4 in \cite{CompOT}). Such regularization does not make the problem convex\footnote{and hence no global minima are guaranteed when solving the problem via projected gradient descent proposed in \cite{Peyre} - see Remark 3 of \cite{Peyre}}, but allows for efficient implementations (Python toolboxes \cite{POT} and \cite{CuturiOTbox}) making the regularization technique widely used for various GW related problems  \cite{Xu2019, Chen2020, Le, Rioux}.

We note that the value of a GW distance is analytically determined in several special cases of metric-measure spaces with infinitely many points: see Theorem 4.1 \cite{Delon} for GW distance between Gaussians when restricting the problem to Gaussian plans, and Theorem 1 of \cite{Arya} for GW distance between spheres. For finite spaces, \cite{Memoliultra} proposes a polynomial time algorithm to compute an ultrametric variant of the GW distance with exponent $p=\infty$ (see Section 5.1 of the reference). We are not aware of a polynomial time algorithm that would solve a general instance of the $p$-GW distance, $p \in [0,\infty)$, between two datasets with finitely many points.

The remainder of this note is organized as follows: Definition \ref{def:GW} defines a GW distance as an optimization program, followed by Example \ref{ex:GW} that illustrates the structure of the program and its non-convexity on a pair of small-size exemplar metric spaces. Theorem \ref{thm:GW} (main result) shows that GW distance between any two metric-measure spaces with finitely many points is a non-convex quadratic program. The result of the theorem is applied in Remark \ref{obs:QAP} and illustrated on the real life data in Example \ref{ex:additional}. The datasets and codes to reproduce Example \ref{ex:additional} are available at \url{https://github.com/kravtsova2/GW_computational}.

\begin{defi}[\textbf{GW distance (finite spaces case), Definition 5.7 of \cite{Memoli}}] \label{def:GW}
Given two finite metric-measure spaces $\mathcal{X}:=(X,d_X,\mu_X)$ and $\mathcal{Y}:=(Y,d_Y,\mu_Y)$ with points $X=\{x_1,\hdots,x_m \}$ and $\{y_1,\hdots,y_n \}$ respectively, define $GW_p^p(\mathcal{X},\mathcal{Y})$, $p \in [1,\infty)$, by the optimal value of the optimization problem\footnote{see also equation (4) of \cite{Peyre} and/or equations (1) and (2) of \cite{Scetbon} that use the same notation as we use here.}
\begin{equation}\label{eq:GW}
\min_{\mu \in \mathcal{C}(\mu_X,\mu_Y)}  \sum_{i,j,k,l} (\Gamma_p)_{ijkl} \cdot \mu_{ij} \mu_{kl} 
\end{equation}
where 
\begin{itemize}
\item The objective coefficients are stored in a 4-way tensor $\Gamma_p$ with elements $(\Gamma_p)_{ijkl} = \left| d_X(x_i,x_k) - d_Y(y_j,y_l)  \right|^p$, $i,k \in \{1,\hdots,m \}$, $j,l \in \{1,\hdots,n \}$.
\item The constraint set $\mathcal{C}(\mu_X,\mu_Y)$ denotes the set of couplings between $\mu_X$ and $\mu_Y$, i.e. Borel probability measures on $X \times Y$ with marginals $\mu_X$ and $\mu_Y$, respectively. 
\item The optimization variables $\mu_{i'j'}, i' \in \{1,\hdots, m \}, j' \in \{1,\hdots, n \}$ stored in a vector representing a probability measure $\mu$ satisfying the coupling constraints above.
\end{itemize}
The p-GW distance between spaces $\mathcal{X}$ and $\mathcal{Y}$ is then given by $\frac{1}{2}\left(GW_p^p(\mathcal{X},\mathcal{Y}) \right)^{1/p}$\footnote{Note that neither taking the $p$th root nor subsequent scaling by $\frac{1}{2}$ affects the nature of the optimization problem; hence, we will be focusing on the object $GW_p^p(\mathcal{X},\mathcal{Y})$ in what follows.}.
\end{defi}

\begin{rema}
Observe that (see Example \ref{ex:GW} for a concrete illustration of these quantities)
\begin{enumerate}
\item The optimization variable $\mu$ can be written as a long vector (of length $m\cdot n$) with entries $\mu_{i'j'}$. 

\item The tensor $\Gamma_p$ can be written in a matrix form by writing the $(i,k)$th block of $\underset{m\cdot n\times m\cdot n}{\Gamma_p}$ to be the $n \times n$ matrix with entries $\left| d_X(x_i,x_k) - d_Y(y_j,y_l)  \right|^p$. Note that this matrix is symmetric due to symmetry of the distances $d_X$ and $d_Y$\footnote{The symmetry is not guaranteed if the distances on the ground spaces are replaced by more general quantities such as asymmetric weight functions on networks \cite{Chowdhury}.}.

\item The constraints $\mu \in \mathcal{C}(\mu_X,\mu_Y)$ can be written in a matrix form $A\mu=\begin{pmatrix} \mu_X \\ \mu_Y \end{pmatrix}$ for a matrix $A$ responsible for summing appropriate entries of the joint probability vector $\mu$. 
\end{enumerate}

 This allows to write the $GW_p^p(\mathcal{X},\mathcal{Y})$ program (\ref{eq:GW}) as a linearly constrained quadratic optimization program 
\begin{align} \label{eq:GW_op}
\min_{\mu \geq 0} &\text{ } \mu' \Gamma_p \mu \\
&A\mu=b \nonumber
\end{align}

\end{rema}

\begin{figure}[h]
\centering
\includegraphics[width=1\textwidth]  {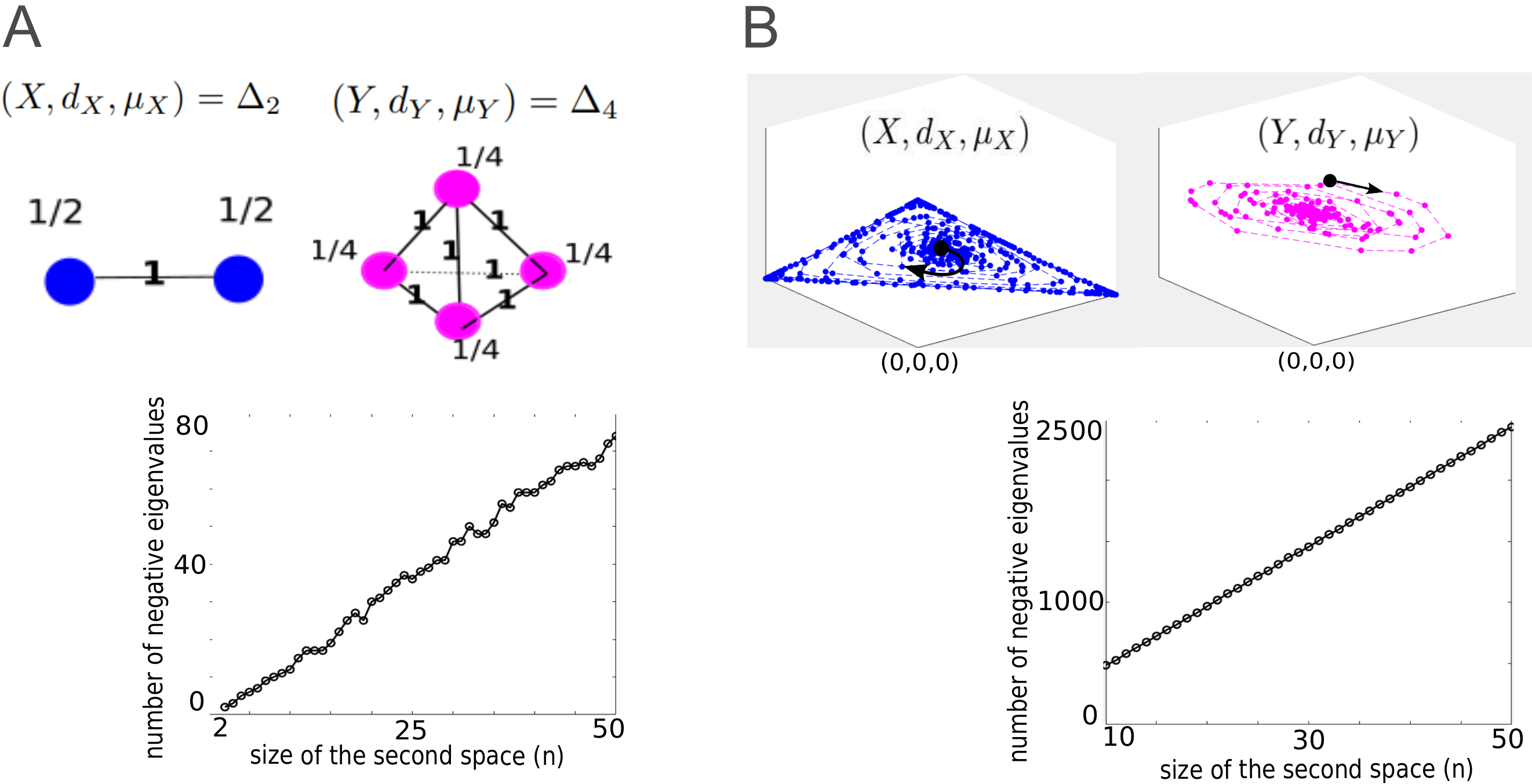}
\caption{\footnotesize{Two additional examples illustrating the presence of negative eigenvalues in $\Gamma_p$, as described in Example \ref{ex:additional}. A. Top panel: a pair of spaces from the family constructed by \cite{Memoli} (Example 5.2). Bottom panel: Fixing the first space with $m=2$ points and  varying the number of points $n$ of the second space results in the $\Gamma_p$, $p=1$, having increasing number of negative eigenvalues. B. Top panel: a pair of spaces constructed in \cite{Kravtsova} using the model from \cite{Xiao} ($m=n=500$ points in each space is shown). Bottom panel: Fixing the first space with $m=50$ and varying number of points in the second space results in $\Gamma_p$, $p=1$, having increasing number of negative eigenvalues.} }
\label{fig:GW}
\end{figure}

The following Example \ref{ex:GW} illustrates the structure of the program in (\ref{eq:GW_op}): 

\begin{exam} \label{ex:GW} (Spaces are from Example 5.1 of \cite{Memoli})). Consider the space $\mathcal{X}:=(X,d,\mu_X)$ with two points $X:=\{x_1,x_2 \}$ with distance $d$ giving $(x_1,x_2) = d(x_2,x_1) = 1$ and $d(x_1,x_1) = d(x_2,x_2) = 0$ (this can be conveniently represented by a distance matrix $(d_{ik})=\begin{pmatrix} 0 & 1 \\ 1 & 0 \end{pmatrix}$) and measure $\mu_X:=(\frac{1}{2},\frac{1}{2})$. A pictorial representation of such space is provided in Figure \ref{fig:GW}A, top panel. Similarly, consider the space $\mathcal{Y}:=(Y,h,\mu_Y)$ with points $Y:=(y_1,y_2)$, distance $(h_{jl})=\begin{pmatrix} 0 & 1 \\ 1 & 0 \end{pmatrix}$ and measure $\mu_Y:=(\frac{1}{4},\frac{3}{4})$. 

\textbf{Note: } The space $\mathcal{X}=\Delta_2$ is a member of the family of spaces defined in Example 5.2 of \cite{Memoli} as $(\Delta_n)_{n \in \mathbb{N}}$, where each member $\Delta_n$ contains $n$ points and is equipped with the distance $(d_n)_{ij} = 1- \delta_{i,j}$ and the uniform measure $\mu_n = (\frac{1}{n},\cdots,\frac{1}{n})$ (the second space $\mathcal{Y}$ above differs in having non-uniform measure). The setting described here is generalized in Example \ref{ex:additional}(a) where $\Delta_2$ is compared with $\Delta_n$ for $n=2,\hdots,50$.

The GW distance seeks to find the joint probability measure with marginals $\mu_X$ and $\mu_Y$, i.e. the vector $\mu:=\begin{pmatrix} \mu_{11} \\ \mu_{12} \\ \mu_{21} \\ \mu_{22} \end{pmatrix}$ satisfying the constraints 
$$\underbrace{\begin{pmatrix} 1 & 1 & 0& 0 \\ 0& 0& 1& 1\\ 1 & 0 & 1 & 0 \\ 0 & 1 & 0 & 1\end {pmatrix}}_{A} \underbrace{\begin{pmatrix} \mu_{11} \\ \mu_{12} \\ \mu_{21} \\ \mu_{22} \end{pmatrix}}_{\mu} = \underbrace{\begin{pmatrix} 1/2 \\ 1/2 \\ 1/4 \\ 3/4 \end{pmatrix}}_{b}$$
which minimizes the objective 
$$\mu' \Gamma \mu = \sum_{i,j,k,l} \Gamma_{ijkl} \mu_{ik}\mu_{jl}$$ where $\Gamma_{ijkl} = |d_{ik} - h_{jl}|^p$. The objective here can be written (in this specific case, it is the same for all $p \in [1,\infty)$) as

\begin{equation}\label{eq:eq:quadratic_form}\underbrace{\begin{pmatrix} \mu_{11} & \mu_{12} & \mu_{21} & \mu_{22}\end{pmatrix}}_{\mu'
} \underbrace{ \begin{pmatrix}|d_{11} - h_{11}| & |d_{11} - h_{12}| & |d_{12} - h_{11}| & |d_{12} - h_{12}| \\ |d_{11} - h_{21}| & |d_{11} - h_{22}| & |d_{12} - h_{21}| & |d_{12} - h_{22}| \\ |d_{21} - h_{11}| & |d_{21} - h_{12}| & |d_{22} - h_{11}| & |d_{22} - h_{12}| \\ |d_{21} - h_{21}| & |d_{21} - h_{22}| & |d_{22} - h_{21}| & |d_{22} - h_{22}|\end{pmatrix}}_{\Gamma} \underbrace{\begin{pmatrix} \mu_{11} \\ \mu_{12} \\ \mu_{21} \\ \mu_{22} \end{pmatrix}}_{\mu}
\end{equation}
The GW distance between $\mathcal{X}$ and $\mathcal{Y}$ is $\frac{1}{2} \left(GW_p^p(\mathcal{X},\mathcal{Y})\right)^{1/p}$, where $GW_p^p(\mathcal{X}, \mathcal{Y})$ is the optimal value of the quadratic program
\begin{align*}
&\min_{\mu \geq 0} \mu' \Gamma \mu \\
&A\mu = b
\end{align*}

For the spaces $\mathcal{X}$ and $
\mathcal{Y}$ defined above, the matrix $\Gamma$ reads
$$\Gamma = \begin{pmatrix} 0 & 1 & 1 & 0 \\ 1 & 0 & 0 & 1 \\ 1 & 0 & 0 & 1 \\ 0 & 1 & 1 & 0 \end{pmatrix}$$

Observe that the matrix $\Gamma$ has a negative eigenvalue $\lambda = -2$, and hence it is not positive semidefinite.  
\end{exam}

Theorem \ref{thm:GW} below shows that the presence of negative eigenvalues in the GW objective matrix is not specific to Example \ref{ex:GW}, but rather holds in general when posing GW problem between any pair of metric-measure spaces with finitely many points. The result states that the matrix in the objective function of GW distance from Definition \ref{def:GW} cannot be positive semidefinite, regardless of an input data. Hence, the resulting quadratic program belongs to a class of non-convex quadratic programs \cite{Vavasis2009}.

\begin{theo}[\textbf{Non-convex quadratic nature of GW distance between finite spaces}] \label{thm:GW} The GW distance program in Definition \ref{def:GW} is a non-convex quadratic optimization program.

\end{theo}
\begin{proof}
Observe that (taking $p=1$ for simplicity), for any general pair of finite metric measures spaces with at least two points and distance matrices $(d_{ik})$ and $(h_{jl})$, a matrix $\underset{m\cdot n\times m\cdot n}{\Gamma}$ as in the objective  (\ref{eq:GW_op}) will always have a submatrix  $\Gamma_{[1:2,1:2]} = \begin{pmatrix}|d_{11} - h_{11}| & |d_{11} - h_{12}| \\ |d_{11} - h_{21}| & |d_{11} - h_{22}| \end{pmatrix}$ in the left corner, i.e. the matrix $\Gamma$ is given by
$$\Gamma = \begin{pmatrix}|d_{11} - h_{11}| & |d_{11} - h_{12}| & \cdots \\ |d_{11} - h_{21}| & |d_{11} - h_{22}| & \cdots\\ \vdots & \vdots & \ddots \end{pmatrix}$$
Since the $(d_{ij})$ and $(h_{ij})$ are the distances, it holds that $d_{11}=h_{11}=h_{22}=0$, while $h_{12}=h_{21}>0$. Hence, the determinant 
$$\text{det}(\Gamma_{[1:2,1:2]}) = 0 - h_{12}h_{21} < 0$$
i.e. $\Gamma$ has a negative \textit{principal minor} (the determinant of a submatrix taken from $\Gamma$ according to index subsets $I \subseteq [m\cdot n]$ and $J \subseteq [m\cdot n]$ with $I=J$). Note that the conclusion would not change if $p>1$, in which case one would still have $\text{det}(\Gamma_{p_{[1:2,1:2]}}) = 0 - h_{12}^ph_{21}^p < 0$.

By \textit{Sylvester's criterion} stating that all principal minors of a real symmetric positive semidefinite matrix must be non-negative (see, e.g., \cite{Prussing}), the matrix $\Gamma$ - or, more generally, $\Gamma_p$, - cannot be positive semidefinite. By the discussion preceding the theorem, a quadratic program with such objective cannot be convex.
\end{proof}

Theorem \ref{thm:GW} indicates that every instance of the GW distance optimization program is non-convex quadratic, which supports the ongoing efforts on its computation (see, e.g., \cite{Memoliultra} for a recent review). At the same time, GW distance can itself be viewed as a relaxation of the Gromov-Hausdorff distance problem (see Section 4.1 of \cite{Memoli}) that generalizes a Quadratic Assignment Problem (QAP) (see Remark 4.6 of \cite{Memoli} for the connection between Gromov-Hausdorff and QAP). Below we discuss the interpretation of GW as a relaxation of QAP in the form that QAP was introduced in the economic literature, and use Theorem \ref{thm:GW} to argue that such relaxation is not convex.

\begin{rema}[\textbf{Relation to the Quadratic Assignment Problem}]\label{obs:QAP} The structure of the problem (\ref{eq:GW_op}) is related to the (NP-hard) QAP introduced by T. Koopmans and M. Beckmann in \cite{Koopmans}. The program describes an optimal assignment of $m$ industrial plants to $m$ locations such that the plants with high flows of commodities between them are placed close to each other (see Section 6 of \cite{Koopmans} for the description). The flows between pairs of plants are represented by a square matrix $(f_{ik})$, and the distances between target locations are stored in a square matrix $(h_{jl})$. The problem is to find a $m\times m$ permutation matrix $X_{i'j'}$ with entries $X_{i'j'} = 1$ if the $i'$th plant is placed into the $j'$th location, and zero otherwise, which describes an optimal assignment of plants to locations. This problem reads
\begin{align*}
& \min_X \sum_{ijkl} f_{ik} h_{jl} X_{ij} X_{kl} \\
& \sum_{j'} X_{i'j'} = 1 \text{ } \forall i'=1, \hdots, n \\
& \sum_{i'} X_{i'j'} = 1 \text{ } \forall j'=1, \hdots, n \\
& X_{i'j'} \in \{0,1 \}
\end{align*}
Writing $X$ as a long vector with $m^2$ entries $X_{i'j'}$ allows to represent the QAP in the form 
\begin{align}\label{eq:qap}
\min_{x \in \{0,1 \}} & x' \Gamma x \\
& Ax = b
\end{align}
with $\Gamma_{ijkl}=f_{ik}h_{jl}$ (see \cite{Bazaraa} for this formulation of QAP).

To see the relation between QAP and GW, consider the objective of the GW distance problem in Definition \ref{def:GW} between metric-measure spaces with ground distances $(d_{ij})$ and $(h_{ij})$, under the choice of exponent $p=2$, which reads 
$$ \mu' \Gamma_p \mu =  \sum_{ijkl} (d_{ik} - h_{jl})^2 \cdot \mu_{ij}\mu_{kl} $$
$$= \underbrace{\sum_{ijkl}d_{ik}^2 \cdot \mu_{ij}\mu_{kl} + \sum_{ijkl}h_{jl}^2 \cdot \mu_{ij}\mu_{kl}}_{\text{depend on $\mu_X$ and $\mu_Y$ only, and thus are irrelevant for optimization}} - \sum_{ijkl}2d_{ik}h_{jl} \cdot \mu_{ij}\mu_{kl}$$
$$=\text{constant} + 2\sum_{ijkl}\underbrace{(-d_{ik})}_{\text{call } f_{ik}}h_{jl} \cdot \mu_{ij}\mu_{kl}$$
The expression above is exactly the QAP objective in (\ref{eq:qap}) if one chooses the flow function $f$ between industrial plants to be the negative distance between them. The optimization variables $\mu$ in GW program are constrained more ``softly" to be in the interval $[0,1]$, in contrast to the ``hard" QAP constraint for $x$ to take integer values in $\{0,1\}$.

This allows to interpret GW problem as a relaxation of the QAP in a framework conceptually similar to interpreting a Kantorovich problem as a relaxation of the Linear Assignment Problem (LAP) (discussed in Sections 2.2 - 2.3 of \cite{CompOT})\footnote{L. Kantorovich and T. Koopmans shared Nobel Prize in Economics in 1975 for their contributions to optimal resource allocation problems \url{https://www.nobelprize.org/prizes/economic-sciences/1975/press-release/}}. As can be seen from Theorem \ref{thm:GW}, the GW relaxation of the QAP is not convex.

\end{rema}

We conclude this note with two additional illustrations of the presence of negative eigenvalues in the GW objective matrix, which demonstrates the conclusion of Theorem \ref{thm:GW} on synthetic and real world data instances (Example \ref{ex:additional} (a) and (b), respectively).

\begin{exam}[\textbf{Two additional illustrations for negative eigenvalues of $\Gamma_p$}, Figure \ref{fig:GW}]\label{ex:additional}
\text{ }
\begin{itemize}
\item[\textbf{(a)}] \textbf{Spaces from Example 5.2 of \cite{Memoli}} This illustration uses the family of spaces $(\Delta_n)$ from Example 5.2 of \cite{Memoli} and described in Example \ref{ex:GW} of this note. Figure \ref{fig:GW}A, top panel, depicts a sample pair $(X,d_X,\mu_X) = \Delta_2$ and $(Y,d_Y,\mu_Y) = \Delta_4$.

To illustrate a presence of negative eigenvalues in the GW objective matrix $\Gamma_p$, we fix the first space to be $\Delta_2$, and vary the second space $\Delta_n$ by increasing the number of points, $n=2,\hdots,50$. For $p=1$, we report the number of negative eigenvalues in $\Gamma_p$, which appears to increase roughly linearly with $n$ (Figure \ref{fig:GW}A, bottom panel).

\item[\textbf{(b)}] \textbf{Spaces based on \cite{Xiao} (see \cite{Kravtsova} for the GW based comparison formulation)} Here we consider pairs of parameterized curves in $3D$ whose points are given by $f(t):=(f_1(t),f_2(t),f_3(t))$ and $g(t):=(g_1(t),g_2(t),g_3(t))$, $t \in [0,T]$, respectively. The functions $f$ and $g$ are taken to be the solutions of the system of tree coupled differential equations constructed in \cite{Xiao} (see equation (6) in the reference) corresponding to two different parameter regimes making the equilibrium point (1,1,1) either unstable or stable focus, respectively (Figure \ref{fig:GW}B, top panel). The problem was considered in \cite{Kravtsova} as a GW based comparison problem (see Figure 3 and corresponding description in the reference), where one views each trajectory as a metric measure space under the arc length distance and a uniform measure on each space.

We fix the first space $(X,d_X,\mu_X)$ to have $m=50$ sample points from the first trajectory, and vary the second space $(Y,d_Y,\mu_Y)$ by sampling $n=10,11,\hdots,50$ points from the second trajectory. For $p=1$, we report the number of negative eigenvalues in $\Gamma_p$, which appears to increase roughly linearly with $n$ (Figure \ref{fig:GW}B, bottom panel).

\end{itemize}

\end{exam}

\vspace{0.5 cm}
\par{\textbf{Acknowledgements and Funding.} The author thanks anonymous referees for pointing out deficiencies in the earlier version of the note. The author thanks Facundo M{\'e}moli for suggesting this problem and supporting the author with NSF DMS \#2301359 and NSF CCS \#1839356. The author thanks Adriana Dawes for the advice and support. A version of the problem considered in this note was given as an assignment during the topics course \textit{Optimal Transport based methods in Data Science} (Math8610) taught by Facundo M{\'e}moli at The Ohio State University in the Spring 2023 semester (\url{https://github.com/ndag/OT-DS}). The author is solely responsible for all mistakes.

\bibliographystyle{plain}
\bibliography{samplebib.bib}

\end{document}